# "What's going to happen after I'm gone?": Parent Perspectives on Technology in Supporting Independent Living for Adults with Intellectual Disabilities

Alexander Tyshka*
Oakland University
Rochester Hills Michigan USA
atyshka@oakland.edu

Andrea Macklem-Zabel*
Oakland University
Rochester Hills Michigan USA
macklemzabel@oakland.edu

Absalat Getachew
Oakland University
Rochester Hills Michigan USA
absalatgetachew@oakland.edu

Foong Ling Chen
Oakland University
Rochester Hills Michigan USA
foongling.chen@gmail.com

Wing-Yue Geoffrey Louie
Oakland University
Rochester Hills Michigan USA
louie@oakland.edu

## Abstract

Adults with intellectual and developmental disabilities (IDD) are increasingly transitioning from family homes towards semi-independent living. As parents hand off the role of primary caregiver, they face numerous challenges in arranging consistent and quality support. Our research centers on understanding these caregiving routines. By focusing on the unique lived experience of parents, who possess extensive explicit and tacit knowledge of their adult child's requirements, we aim to map the management of care they provide. This foundational understanding is essential to identifying how assistive technologies can effectively serve a key role in supporting adults with IDD in this transition. In this work, we interviewed 16 parents of adults with IDD beginning this transition to understand: 1) how they currently provide support for daily living and what makes their support effective; 2) what their experiences and perceptions are regarding the use of technology; and 3) how they envision assistive technology successfully integrating with their adult child's new home or with other supports to promote independence. Our thematic analysis produced four themes: common modes of support, inside the routine to support growth, the fragility of continuity of care across transitions, and participants' perceptions and experiences of assistive technology. Building on these findings, we derive four design principles for growth-oriented assistive technology. These include pre-transition onboarding to capture caregiver tacit knowledge; structured scaffolding towards long-term growth; adaptive sensing that responds to day-to-day variability; and customization for the individual balanced with consistency for the care network.

*Both authors contributed equally to this research.

This work was supported by the National Science Foundation CAREER award #2238088 and the Oakland University Community Changemaker Challenge Grant.

## 1 Introduction

For many adults with intellectual and developmental disabilities (IDD), the desire to live in their own home runs deep. It reflects the same drive for autonomy and social connection they see in their neurotypical peers [66]. This is accompanied by practical urgency as improved care has led to individuals with IDD living longer, often well beyond the age their parents can continue providing daily support [52]. As a result, there has been a growing movement in the IDD community towards exploring long-term housing solutions such as semi-independent living [7].

Arranging housing is only part of what parents are managing during this transition. Parents of adults with IDD are not simply caregivers but are the primary holders of deeply personal and accumulated knowledge of their child. This is built through decades of daily interaction and has been described as tacit knowledge. Tacit knowledge enables the ability to read subtle behavioral signs, predict when something may go wrong, and identify what a child needs in ways that cannot easily be handed off [41, 55]. This knowledge shapes daily support in ways such as determining what communication cues to use, when to initiate a task, how to adapt to meet a child where they are that day, and when to step in versus step back. For example, a parent may know their child needs a second verbal cue before starting laundry, or that a difficult morning means the medication routine will require more time. Yet when parents are no longer the primary caregiver, this knowledge does not carry forward and skills that were stable at home can deteriorate rapidly in new settings [31, 60].

Technology has potential to play a key role in carrying that support forward as parents step back. Assistive technology is already embedded in the daily lives of many individuals with IDD, supporting time management [2], task completion [27, 44], navigation [35, 64], and remote access to support staff [69, 78]. For families approaching a transition to semi-independent living, this existing technology presence matters.

Yet what current assistive technology supports and what the transition requires are not aligned. Prior systems have supported task performance in specific settings such as transportation [13], workplaces [24, 37–39], school [79], and home routines [40]. Most technology is designed for task completion in the moment rather than growth. It prompts, reminds, and assists but does not adapt as skills develop or carry forward what works for a person across caregivers and settings [20, 32]. What these systems fail to capture is the tacit knowledge parents rely on to make growth-oriented support effective across the transitions in an individual's life.

Addressing this gap requires understanding perspectives of those who hold that knowledge. Research has examined how individuals with IDD experience technology [2, 33] and how support staff perceive and adopt it [58, 63], yet parents and family caregivers have only been consulted about reactions to specific technologies rather than the broader expertise they hold on what makes support effective. This gap is significant because in semi-independent living contexts, parents often advocate for and implement assistive technology [51]. The transition to semi-independent living also presents a rare opportunity to capture this perspective at an inflection point where families are actively rethinking what support looks like, what can be maintained, and what might be carried forward by technology. Capturing their perspective here can reveal not just what technology should do, but what it needs to understand about the nature of effective support over time.

In this work, rather than centering our research on a specific technology, we center it on the routines of caregiving. We investigate how tacit knowledge of parental caregivers can inform the design of assistive technology that aims to support the transition to semi-independent living.

We seek to answer the following research questions:

*RQ1:* How do parents currently provide support for daily living? What makes support effective?

*RQ2:* What are their experiences and perceptions regarding the use of technology?

*RQ3:* How do they envision assistive technology integrating with their child's new home and with other supports to promote independence, and what will it need to succeed?

To answer these questions, we interviewed parents of adults with IDD who are transitioning towards semi-independent living. Through thematic analysis, we surfaced the accumulated tacit knowledge that underlies effective parental support, strategies through which that knowledge is enacted in daily life, and fragility of that knowledge at the moment of transition.

## 2 Related Works

Our work builds on two bodies of literature: 1) research on parental tacit knowledge and stakeholder perspectives on assistive technology, and 2) the landscape of assistive technologies for independent living. These establish both why the parent perspective represents a critical design resource and how existing tools have not yet been designed to carry that knowledge forward.

### 2.1 Parental Tacit Knowledge and Stakeholder Perspectives on Assistive Technology

Parents of adults with IDD occupy a distinct position in the support ecosystem. Unlike paid staff who rotate across clients, parents have spent decades in daily proximity with their child and developed an understanding that goes beyond procedural knowledge [6, 28, 29]. Research has described this as tacit knowledge: a form of implicit, embodied knowing built through repeated interaction and shared history that cannot easily be made explicit or transferred to others [41, 55]. In the context of IDD care, this tacit knowledge manifests as the capacity to read subtle behavioral signals, anticipate needs before they are expressed, and calibrate support in ways that are deeply specific to the individual [6, 41]. This knowledge is not a fixed set of facts about their child, but a skill built through years of reading and responding to them. Because this knowledge is embodied and relational rather than documented, it does not easily survive changes in the caregiving context [31, 41]. The transition to new settings introduces significant challenges because support staff rarely have access to the depth of understanding that parents have built over years [60]. What is lost in these transitions is the practical judgment that makes support effective for a specific person. Without it, skills that were stable at home can deteriorate in new settings [31, 60], precisely when the individual is adjusting to a new environment and new routines.

For assistive technology to serve this transition, it would need to do something no current tool has been shown to do: capture and carry forward the kind of individualized, experience-based knowledge that parents hold. Yet the assistive technology literature has not examined this question from the perspective of parents. Research has explored how individuals with IDD perceive and use technology [2, 33] and how support staff experience adoption and implementation [58, 63]. Individuals with IDD generally hold positive attitudes toward mobile devices and computing technology, though usage of assistive features for daily routines remains notably lower than for social and entertainment purposes [57]. Technology for adults with Down Syndrome has been studied more closely, with findings showing that positive reinforcement, ability recognition, multimodal interaction, adaptability, and personalization are central to sustained engagement [33]. Staff perspectives reveal that practical barriers around setup time and system complexity undermine use, particularly when there is staff turnover [63]. The studies that do include family voices tend to capture reactions to specific technologies [10, 51, 65, 69, 77] or how families adopt and abandon particular tools [17, 18, 76]. This leaves parental expertise unexamined as a design resource because studies only capture whether a tool works for their child with IDD and not the knowledge that makes support effective in the first place.

### 2.2 Assistive Technologies for Independent Living

To understand the current state of assistive technology for individuals with IDD, we review what has been developed, what has been shown to work, and where gaps remain in carrying forward the individualized knowledge that makes support effective through the transition to semi-independent living.

*2.2.1 Mobile Apps and Electronic Planners.* Mobile technologies are the most widely adopted assistive technologies among individuals with IDD because of their accessibility and low cost. Electronic planning applications have demonstrated effectiveness for delivering reminders and step-by-step instructions for activities of daily living via pictures and audio/video [27, 68] and increasing self-determination [67]. Research has also explored integrating mobile prompts with environmental triggers to situate support within the physical environment [12, 44]. Desideri et al. [20] found very limited work on intelligent systems that can sense context and provide prompts in real time. Instead, most apps operate on fixed schedules and predetermined prompts that deliver the same support regardless of the individual's state on a given day or how their abilities have changed over time. The prompt structure must be manually reconfigured by a caregiver when support needs shift, which places the burden of adaptation entirely on the human [20].

*2.2.2 Voice Assistants and Smart Speakers.* Voice assistants and smart speakers are widely used among individuals with IDD for reminders, information retrieval, and daily routine support [22, 33, 65]. These conversational interfaces remove the need to navigate complex visual interfaces and makes them accessible to individuals with limited literacy or fine motor challenges. However, these tools are designed for neurotypical speech patterns and lack features for accommodating non-standard communication styles [72]. Rather than adapting to the individual, the individual must adapt to the technology. Beyond communication, voice assistants treat each interaction as largely independent and do not accumulate the history of a user's needs, struggles, or progress.

*2.2.3 Smart Homes, Ambient Assisted Living, and Remote Support.* Smart homes are well-positioned to address the gap in adaptive context-aware prompting, though far less research has explored their application for individuals with IDD compared to older adult care [20, 48, 59]. Landuran et al. [46] evaluated the DomAssist platform in a long-term deployment for adults with Down Syndrome, finding high usability and increased self-determination. Lancioni et al. [42, 45] similarly demonstrated that tying instructions to actual task performance could support independence. Remote support technologies extend monitoring further by enabling staff to observe and respond in real time [19, 69], but like smart home systems, they arrive at each interaction without a model of the individual built over time. This misses detecting growth, regression, or anomalies in a specific person's behavior.

*2.2.4 Assistive Robots.* Assistive robots are an emerging technology for supporting independent living. Wolbring and Yumakulov [74] surveyed disability care staff and found most preferred robots to handle physical tasks, emphasizing that robots should do activities *with* people rather than *for* them. Xu et al. [77] and Kong et al. [36] found that families and adults with IDD held positive views of social robots and expressed interest in expanding their role to support daily living and safety. Van Dam et al. [71] found that reminder robots were helpful for maintaining routines. However, a recent survey [53] found that AI-equipped robots were the most studied technology, with the vast majority focused on teaching social skills for children with autism, leaving daily living support for adults with IDD largely unaddressed.

*Summary.* Across this literature, a clear gap emerges. Parents hold the knowledge that makes support effective, but it has not been studied as a resource for design. Meanwhile, assistive technology remains focused on task completion in a single setting and not on growth over time [20, 32] or on transitions in the living situation where knowledge about the adult with IDD is most at risk of being lost. This study addresses that gap by asking what parents of adults with IDD know, how they apply it, and what it would take for technology to carry it forward.

# 3 Methodology

We conducted a semi-structured interview with parents of adults with IDD to surface the tacit knowledge parents have accumulated through years of caregiving, their experiences with assistive technology, and their vision for how technology could support their adult child's transition to semi-independent living.

## 3.1 Participants

A total of 16 participants were recruited through a local organization that is developing semi-independent housing communities for adults with IDD, while two participants were recruited through referrals. The inclusion criteria for participation were as follows:

- The participant is a parent of an adult with IDD
- Their adult child with IDD is preparing to transition to semi-independent living within the next several years (i.e at the beginning of the transition)

Our participants come from 13 distinct families. We interviewed 12 mothers and 4 fathers, ranging in ages from 48 to 78 years (mean = 66.4). The adults with IDD were aged 20–46 years (mean = 32.4); 10 were male and 3 were female. Two of the adults with IDD were living in group homes with frequent visits to their parents and the remaining lived at home. Participants received a $15 gift card for their time and the study was approved by the university's institutional review board.

Parents described their adult children as participating in a range of daily and community activities. A majority were reported to be employed or involved in vocational programs. Parents also reported meaningful variability in communication styles and literacy. Seven individuals were described as having difficulty verbalizing, and six had limited reading proficiency. These differences shaped how parents described providing support in daily life, how accessible existing technologies were to their adult children, and provided important context for the themes that follow.

## 3.2 Interview Protocol

Interviews were conducted primarily online using the Zoom video conferencing platform but participants also had the option to meet in person. Interviews were designed to last one hour but some sessions extended when participants requested additional time (~70 minutes).

The semi-structured interview consisted of four sections designed to progressively surface parental expertise and its implications for assistive technology design. The first section focused on understanding the background of the parent, the adult child with IDD, their relationship, and their plan and motivation for transitioning to semi-independent living. The second section centered on current support systems for participants describing where, when, and how they provide support in daily life. This captured the accumulated knowledge underlying their caregiving decisions. The third section addressed current and past use of assistive technology and family's preferences for different technologies. The final section explored areas they thought would be most transformative for technology to provide support and increase quality of life.

## 3.3 Data Analysis

We analyzed the interviews using reflexive thematic analysis with Braun and Clarke's six-phase approach [11]. All interviews were audio/video recorded and transcribed by one researcher. Two members of the team then conducted an initial round of inductive coding. Each developed their own set of reflexive codes while engaging in regular discussions to reflect on interpretations and emerging patterns. This process was to explore multiple perspectives on the data rather than to reach agreement. Following initial coding, the two coders met to reflect on their codes and identify candidate themes. These discussions were then expanded to include the full team that read the transcripts to support iterative refinement of the coding structure, themes, and broader interpretation. Based on these reflections, one author re-coded the dataset using a revised coding scheme to ensure consistency across the data and keeping with the reflexive thematic analysis approach. The final analysis resulted in 104 codes organized into four themes.

# 4 Themes

Parents used their unique understanding of their adult child's capabilities, struggles, and support needs to shape decisions that promoted agency and fostered growth. The themes that follow traced the arc of that support. Theme 1 (Common Modes of Support) examined how parents selected ways to help keep their adult child participating and growing in daily activities. Theme 2 (Inside the Routine) explored how parents built and maintained routines that made growth possible. Theme 3 (The Fragility of Continuity Across Transitions) turned to the accumulated knowledge sustaining that support and risks it faces as families approach semi-independent living. Theme 4 (Participants' Perceptions and Experiences with Assistive Technology) captured parents' assessments of existing tools and their vision for what assistive technology could do differently.

## 4.1 Theme 1: Common Modes of Support

Parents drew on common modes of support for their adult child in daily home life to promote agency and foster growth. These modes of support included creating task adaptations, self-adaptation via routine building, and choosing when to do things on their adult child's behalf (Fig. 1). This theme also examines how parents believe technology could strengthen these forms of support.

*4.1.1 Task Adaptations.* When there was a mismatch between competencies and task requirements, one of the supports was to implement an analog or technology-driven solution to simplify task requirements.

Fine motor skills were a common task area, as expressed by P2: *"He doesn't have really good fine motor skills, so we avoid anything with buttons or zippers [...] And he can't tie his shoes. So we have slip-ons."* Parents also augmented text-based interfaces with visual aids to enable independence in tasks. As P6 shared, *"He does his own laundry with probing from me, but I have little marks on where to set the water, where to set the temperature, and he does it totally independently. That's all it took [...], he could see the marks; he maybe couldn't read hot / cold."*

Communication was another common adaptation families implemented with visual or auditory aids for comprehension and self-expression tasks. Parents described a wide variability in their adult children's communicative repertoires that included verbal expression, gestures, picture boards, text, and partial literacy. Physical and digital picture aids were a way some families facilitated communication to support agency. P3 talked about grocery shopping with a picture list. *"I have a picture grocery list for him so that he can identify what he needs. [...] it's a picture of apples, a picture of raspberries, and then I circle each week what he needs to purchase."* Voice-based technologies were another adaptation to make information accessible. P10 shared, *"She uses Alexa to figure out what she's gonna wear the next day. She'll ask her about the weather and all that."* Parents described how their adult child had to adapt to the limits of technology rather than technology adapting to diverse communicative repertoires. P3's son had recently begun experimenting with voice-to-text, *"he'll take his phone and say, 'Siri, send a text to Dad'. He tends to over talk [...] and so Siri gets a little confused because he's giving too much information."* Supporting effective communication required patience and responsiveness, re-prompting to ensure preferences were not lost to recency bias or convenience.

Overall, environmental and user interface adaptations were a preferred means of promoting agency because they made the world more accessible. However, in many scenarios task adaptations hit roadblocks and fell short of producing an interface that addressed the communication needs of an individual.

*4.1.2 Self-Adaptation via Routine Building.* While task adaptations enabled immediate participation in tasks, it was necessary for adults with IDD to learn to also self-adapt. In daily living tasks where high cognitive load or impaired executive function created barriers to independence, routine building served as an enabler for self-adaptation. In the words of P10, *"she has slow processing, and that's an issue. You know she's done so amazingly well for somebody with her very low IQ."* Parents worked over years to build skills and muscle memory through routines for activities of daily living such as laundry, hygiene, cleaning, and food preparation. As P8B expressed, *"you put something in place, and then you just gotta work it and work it and work it[...] it takes a long, long time for him to [...] really start following through [...] day after day".* For some, learned skills could become self-sustaining, but in other cases consistent maintenance of a skill was needed to prevent regression. As P11 shared, "*sometimes he can remember 3 or 4 steps, other times his short-term memory is very poor. If you don't constantly maintain it, he'll lose the skill.*" Parents envisioned how technology might reinforce these routines and reduce caregiver reliance in developing and maintaining skill mastery. P2 imagined voice-guided systems, *"if there could be a voice or something, 'okay, Dennis, let's get your breakfast ready. Let's get the oatmeal out.' Yeah. 'Let's, Add some water to it, stir it up, and put it in the microwave for two minutes.'"*

*4.1.3 Deciding When to Do For.* When fostering growth and agency, parents noted the importance of doing things with their adult child, rather than for them. However, there were times parents felt it necessary to perform functions on behalf of their adult child. Their reasons included current abilities, time limitations, and safety.

Parents performed certain components of tasks to address gaps in their adult child's competencies where agency and growth was not a reasonable goal. One such area was qualitative judgments and fine-grained decision-making. Using driving as an analogy, P10 described how her daughter tended towards binary extremes, *"She's either going with her foot all the way down to the floor on a pedal or not at all, like she doesn't have that subtle... You know how you learn the touch"* P3 demonstrated how this applied in practice: *"We adjust the temperature in the shower for him, and then leave, and he takes care of the rest of it."* In other areas like scheduling, transportation, and financial matters, parents handled the entire task themselves for their adult child. Parents noted that in certain areas, independence could not be fostered due to perceived and/or experienced serious negative consequences. P10 warned, *"we don't let her use the stove because it's hot, and it easily surprises you."*

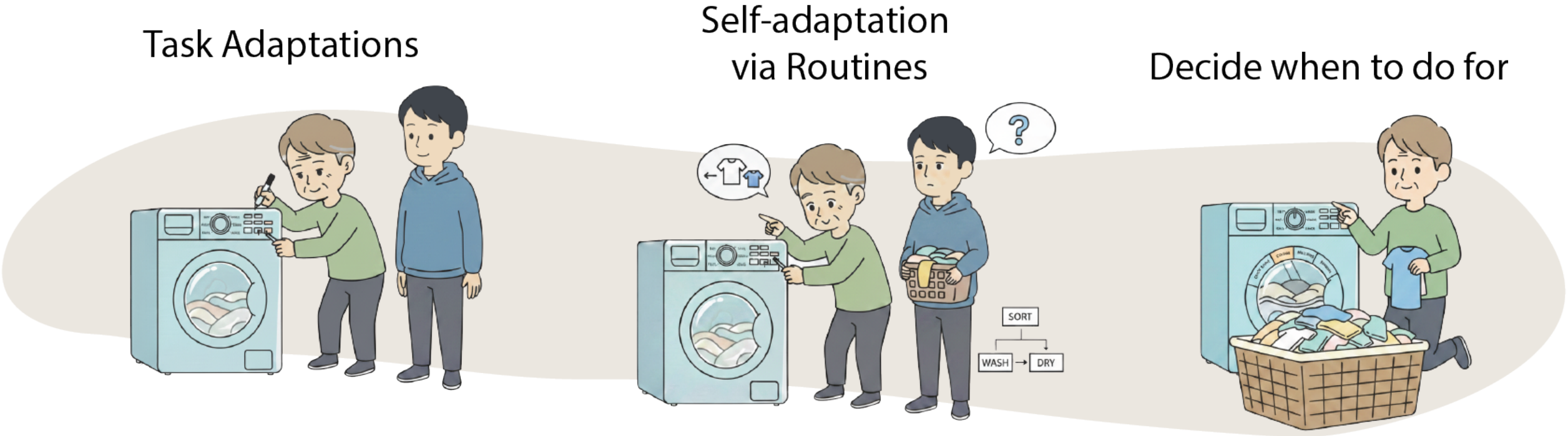

Figure 1: The three common modes of support identified in Theme 1, illustrated through laundry, a chore commonly mentioned across the interviews. The laundry example will be used consistently through the paper to illustrate how a single task can have many complexities. Left: task adaptation, where the parent marks the machine so their adult child with IDD can set the temperature independently. Middle: self-adaptation via routine building, where the parent teaches the steps of washing clothes to grow a new competency. Right: doing-for, where the parent folds the clothes on behalf of their adult child because the skill is not yet developed.

Several participants noted that while their adult child was capable of performing certain tasks on their own or with some assistance, parents took over when time was short. As P5 put it, *"sometimes [it's] easier to do it for them than it is to wait for them to do it themselves. [...] all of us struggle with that balance."* P11 explained that timelines sometimes tipped this balance, *"if you want to get out the door quick. that's when we tend to take over,"* even though they knew it risked stalling growth, *"we really got to figure out his capability [...] and learn for himself."*

While navigating when to intervene, parents acknowledged struggling with their own limiting beliefs or personal preferences about how things should be done. Both of which can unintentionally hold their adult child back. P5 illustrates this tension:

> *"He'll let me support him up to the point that I treat him like a baby. [...] I try not to do that and I don't think I infantilize him. But once in a while he'll express a desire to do things more on his own that surprises me. He's cooked a pizza in the oven when... [laughs] I had no idea he could do it on his own. I think that's a way of him telling me 'I can do this.'"*

*4.1.4 Design Insights.* Current assistive technology for individuals with IDD focuses largely in task adaptation mode by modifying interfaces and environments to lower the barrier to participation [20, 34]. Our findings extend this existing literature to understand: 1) how current task adaptations are put in place; 2) how task adaptations alone are not sufficient and how self-adaptations towards growth is a crucial mode of support; and 3) how self-adaptation is built through routines.

## 4.2 Theme 2: Inside the Routine

Looking deeper into self-adaptation via routine building from Theme 1, a common structure parents described was initiating, assisting, and verifying (IAV) to build independence (Fig. 2). Rather than a static set of actions, this structure served as a form of scaffolding that adapted in relation to their adult child's competency and daily life. Theme 2 examines how parents implemented this practice and what was required for technology to carry it forward.

*4.2.1 An Observed Pattern of Support for Growth.* Initiation of tasks was done through prompting, and the quantity varied across families depending on each individual's ability to initiate tasks independently. Some parents described prompting for nearly every step, while others described individuals who could begin routines on their own but needed occasional reminders to sustain momentum. P5 expressed, *"he won't ask. [...]. One of us [is] initiating his support, but we try to let him do what he can on his own."* External aids such as calendars extended support beyond the parent by cuing the initiation of tasks like laundry and reducing the need for repeated verbal reminders. P13 explains, *"Well, it doesn't work just to say, Daniel, do your laundry every week, [...] So I put out a schedule on a calendar, [...] but also I have to [tell him to] finish his laundry, because he might go start it, and then it might sit there for 3 days."* Other tools served similar functions. As P3 noted, timers provided a mechanism for signaling transitions, *"we set timers, like, you know, in whatever amount of minutes we have to do whatever is next."*

Assistance covered the verbal guidance and physical help parents provided during activities of daily living. How it was enacted was personalized to the individual and the demands of the task. Hygiene and dressing were common areas where verbal guidance and partial physical help were necessary, as steps might be skipped or performed superficially. P5 explained, *"he brushes his own teeth, but I have to tell him every single time, don't just hold the toothbrush into your mouth. And so a lot of support with tooth brushing, flossing."* P7 described more direct physical involvement, *"you know I need to do hand over hand to wash his face and brush his teeth,"* with assistance extending into dressing. P5 and P7 needed more parental involvement. Other parents described their adult child as quite independent, requiring only occasional reminders.

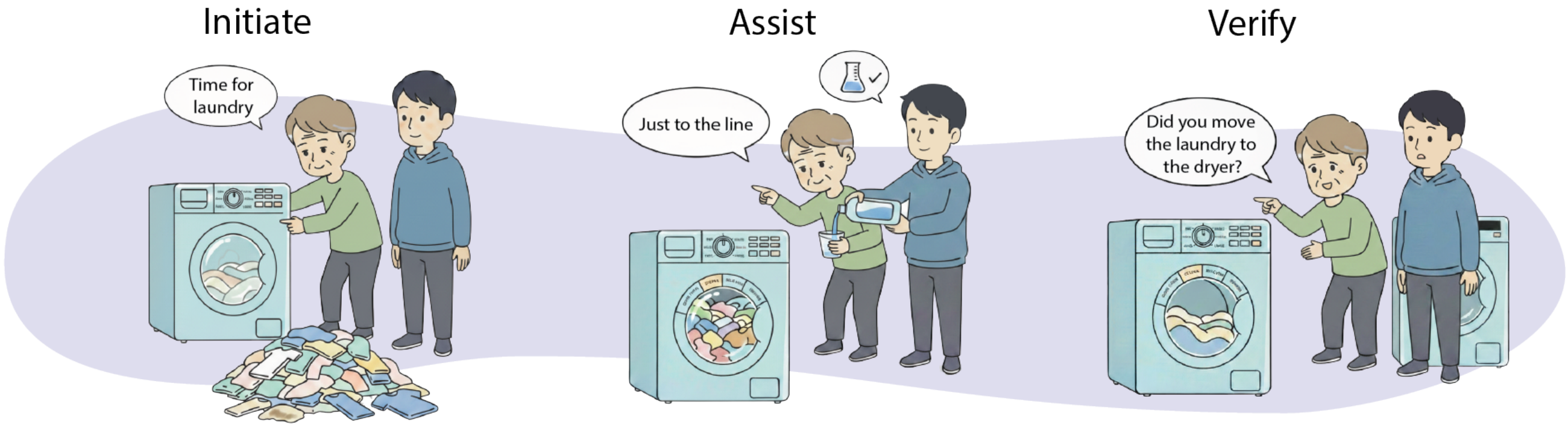

Figure 2: Initiate, assist, verify (IAV) captures the common structure of support identified in Theme 2, illustrated through laundry. Left: initiate, where the parent prompts their adult child with IDD to start the task. Middle: assist, where the parent verbally and physically guides their adult child through the steps. Right: verify, where the parent checks whether their adult child has moved the wet clothes to the dryer.

Verification was how parents ensured tasks were attempted and completed to an acceptable standard while sometimes making small corrections. P12 described checking their daughter's shoes after she tied them, *"she's really just finally focused on knowing how to tie your shoes, you know, making sure those are secure. Sometimes we'll check them and have [to] redo them.*" P5 described a similar pattern, *"so bathing he bathes himself, but I kind of touch it up at the end."* Verification was not simply a quality check but a judgment shaped by accumulated knowledge of what completion should look like for that individual. A new caregiver or assistive technology might inherit the responsibility to check whether a task is done, but without that accumulated judgment, the standard applied risks being generic rather than calibrated to the individual.

*4.2.2 Calibrating Across Two Horizons.* The IAV structure of support was not fixed. Parents calibrated the structure across two timescales: 1) day-to-day based on demands of any given day and 2) long-term based on the adult child's growth goals (Fig. 3). These time horizons meant that support was dynamic, responsive to the moment, and strategically considering the future growth of their adult child.

***Day-to-Day Calibrations.*** Once a support structure for a task was in place, parents described how the conditions of a given day shifted how much and what kind of support was needed. These sources of variability included internal changes in an individual's attention or emotional state and external changes in the environment around the routine.

Internal variability meant that the same task could require different amounts of support depending on the adult with IDD's internal state. Parents described reading their adult child's attention and mood and adjusting accordingly. P11 captured this support recalibration given what an individual could sustain that day, *"we've all got good days, and we've all got bad days. So some days he's great and he's on task and everything else. While, other days he might need [...] more prompting, more verbal, depending on what it is."*

External variability required parents to adjust routines to environmental conditions. Weather was a source of friction, particularly during seasonal transitions when clothing choices could not be left to established routine alone. P3 described, *"I say get a short sleeve shirt, he'll grab a long sleeve shirt. [...] Once we get into a season and those things are [in] the middle of his closet, he can do that completely independently."* External variability meant support decisions are never static. The support structure might be established but active judgment was necessary for the calibration to any given day.

Parents imagined technology a natural fit for providing support based on environmental changes. P3 envisioned a weather display that communicated through images rather than temperatures, *"a weather app [...] 80 degrees, that has no value to him. But if it had some kind of [...] picture, like a bright sunshine meant shorts and a T-shirt and a half a cloud."* Currently, parents bear the responsibility of making contextual information accessible to their adult child. Future assistive technologies must be able to both identify and address how information should be presented to an individual.

***Long-Term Growth Focused Calibrations.*** The scaffold parents built was never meant to be permanent. As their adult child's competence grew, parents described intentionally stepping back by fading prompts, redistributing tasks to tools, and allowing mistakes as a part of building independence.

Parents described support as something that could be reduced or reconfigured as skills developed. P1A explained how hands-on guidance was often necessary for new tasks but could eventually shift to monitoring: *"Let's say he's got a treadmill, [...] when he first got it [he had] to learn how to operate it. So there's hands-on [support]. Once he learns something [...] it's you know, kind of monitor that he didn't fall off the bike."* P7 described intentionally fading the level of assistance with their prompts by shifting from physical to verbal wherever possible, *"We're always working on trying to improve his activities of daily living in terms of what he does; trying to just do*

Calibrating Across Two Horizons

Day to Day Calibrations

Long-term growth focused calibrations

My calendar reminds me every week

Forgot the dryer sheets! Oh well, next time!

Internal variability

External variability

Fading prompts

Redistribute support to tools

Allow for mistakes

Figure 3: IAV calibration categories, illustrated through laundry. Day-to-day calibrations include internal variability, where the adult with IDD starts the task frustrated, and external variability, where a detergent change from liquid to powder causes confusion. Long-term growth calibrations include fading prompts, where the parent steps back as their adult child has learned the skill; redistributing support to tools, where the adult with IDD uses a calendar instead of relying on their parent; and allowing mistakes, where forgetting a dryer sheet is treated as a minor deviation rather than an error.

*a few physical prompts rather than doing it all, fading the physical prompts and doing verbal when we can."*

Tools were imagined as a means of redistributing support away from the parent while reinforcing routine. P12 envisioned how checklists could support her daughter's independence, *"She may need. [...] checklists, because then you're not forgetting a step, but once she understands the steps, she's pretty good. [...] she would not have to rely on someone 24/7."* P13 described already using schedules to lighten their parental prompting, *"so, rather than us, always saying, [...] Did you do this? We can just say, check your schedule."* P2 added that technology might provide both functional support and improve self-confidence. They described how it could work around their adult child's functional limits: *"His fine motor skills that aren't as good. Technology could kind of get around that or, you know, do things in a way that he could do it successfully."* But the bigger gain they saw was confidence: *"With technology [...] he might be able to do [things] even [if it's] simple things, [it's] more [that] him taking care of himself would give him more confidence in living away from us."* P6 reflected broadly on the task opportunities for technology, *"I think the knowing the weather, telling the time. You know, reminders about things.[..] I would think just about any of the areas that we currently support him with. I think technology could serve a really meaningful role for him."*

Allowing mistakes was something parents saw as essential to how their adult children developed independence in performing tasks. P13 reflected, *"if he makes a mistake, who cares? Those are the things you just let him make a mistake or let him do something that's different, that maybe he wants to do it a certain way."* P10 added that mistakes built problem-solving skills that were *"an intangible [thing] that lasts them their whole lives."* P11 extended this further, questioning whether process deviations should be treated as errors when the outcome was acceptable. *"If someone does it in the wrong direction, but ends up with the same end goal. Is that okay? Because I think sometimes we expect it to be done A to Z in the correct manner, and it really doesn't matter."*

*4.2.3 Design Insights.* Current assistive technologies such as reminder/schedule apps and assistive robots support this IAV structure. The critical difference is that they adapt prompts to focus only on task completion, regardless of an individual's state that day or long-term growth focused goals [20, 27, 71]. This theme extends literature by highlighting that for assistive technology to support IAV meaningfully, it needs go beyond only adapting prompts for task completion and be responsive to day-to-day internal/external conditions and track growth/regression over time while considering an individuals with IDD's baseline.

## 4.3 Theme 3: The Fragility of Continuity Across Transitions

Themes 1 and 2 revealed the parents' processes for providing care by identifying the right approach, developing a structure, and adapting it to foster growth. This support was made possible by the knowledge they developed over decades with their adult child. Theme 3 examines: 1) how this built up tacit knowledge enables them to know what to monitor to anticipate their adult children's hidden needs; and 2) what is at stake when knowledge regarding support fails to transfer during transition across caregivers, environments, and systems (Fig. 4).

*4.3.1 Hidden Needs.* Parents described how their support necessitated ongoing vigilance of small shifts and subtle signs that something had changed. Parents have learned to pay attention to silence, absence of signals, and minute behavioral changes to identify needs for support. This knowledge, built from years of experience, could not easily be written down. It represents an accumulated understanding of their adult child's hidden needs. Someone who is less familiar would likely overlook these subtle behavioral cues. This is further complicated by the limited self-advocacy skills that many parents identified as a challenge for their adult child. P10 shared:

> *"If she wants something or needs something, she's not real good about advocating. She'll just sit there and wait for somebody to notice that she needs something.*

> *And that is not gonna work for her when she gets in [a new residence] because the staff will not[...] they're not gonna wait on her hand and foot.*

Issues can go explicitly unexpressed for days, which can be extra concerning when there are health and safety consequences. P13 highlighted a situation where a skill was established and they created room for independence but an incident led to a modification in their support behavior:

> *"But [one day] he decided he wasn't gonna take [his medicine] [...]. He went in for his checkup [...] and the doctor said, 'Oh, my gosh!' [his lab results indicated dangerously high numbers] [...] Those are the kind of things we just gotta kinda keep an oversight on [...] [Now] we're watching to make sure he takes it every day."*

While some families had no concerns about leaving their adult children unsupervised, for others, even brief lapses in monitoring could create profound setbacks in independence. As P7 recounted,

> *"He'll beeline to the brownies and the cake, and he's gluten free, which makes it even more challenging [...] he almost choked to death actually [...] Nobody was in the room because they were getting a utensil for someone else. He ended up on a ventilator for 11 days [...] We had to teach him how to stand and walk again."*

Vigilance monitoring was a proposed future support. Participants imagined technology extending monitoring beyond what they could sustain themselves, particularly as their adult child moved toward semi-independent living. As P7 considered, *"I think a sensor on the floor to know when he's out of bed would be a good thing to have, especially when he's living in an apartment."*

These examples show the importance of monitoring hidden needs. What makes it possible is not vigilance alone but an intimate knowledge of the individual to anticipate needs. Without knowing what normal looks like, there is nothing to interpret. These examples make it clear that support itself is not sufficient for growth because without the knowledge parents built across years of caregiving, growth can become fragile.

*4.3.2 Breakdowns in Knowledge Transfer.* A recurring concern was what would happen when this knowledge is lost. When asked why they were exploring semi-independent living at this time, P4 put it plainly, *"it's the same reason everybody's going to say... what's going to happen after I'm gone."* P5 emphasized the importance of proactive planning, *"it's better for him if he's settled in an independent living situation before there's a medical crisis or death in our family of his parents."* P6 envisioned that technology could be that all important *"game changer"* once caregivers are gone, because *"parents and caregivers aren't always going to be here."*

Parents provided vivid examples of what happens when knowledge does not carry across contexts. Skills that were learned with success at home deteriorated in other contexts because they were not consistently supported. P8B described how even when a communication system was learned successfully, its effectiveness depended on consistent support across contexts:

> *"You know, to learn to use a system or communicative device... you have to, of course, do the upfront training right [...] which means the family has to be on board 100%, right? That hasn't been an issue. But where... what's been an issue is, then when he steps outside, going to the schools, going to work, [...] people just don't use it, [...] And if he's not able to use it in all those environments, then it makes it very difficult to transfer."*

Even when skills have been mastered and well established, regression can occur because support knowledge is not transferred across contexts. As P9A explained, *"he's regressed in a lot of his skills that he had before. Just because he hasn't been in the right kind of supportive environment for using them."* Even seemingly small details, such as a person's handedness, could take months to rediscover when not transferred. P11 recalled, *"It took [the new support staff] 6 months to figure out Harry was left handed. [...] I gave [her] that information at the start of the semester. [...] He's left handed. He has to pick up the left handed scissors."* These examples highlight that what is lost is not just information, but the knowledge that allows others to act effectively on behalf of the person. Without deep knowledge of an adult with IDD, there is a risk that they must start from scratch redeveloping skills with each new caregiver or setting.

Several parents emphasized that technology could help maintain consistency across caregivers if it could retain knowledge, apply it, and support its transition across contexts. P11 envisioned AI that could *"keep problem solving and keep focus on the consistent [...] education of the direct care worker,"* noting that *"transfer of information between people is terrible... there needs to be centralized information. [...] We are not helping him. You know. We are untraining him, or we're not pushing him enough. These are all things that I think could be learned within AI."* In this view, technology could act as a living repository of support knowledge that bridges the gap between caregivers.

*4.3.3 Design Insights.* Current assistive technology is designed to integrate into the established support ecosystems surrounding an adult with IDD. This assumes access to an established caregiver who can draw on their deep tacit knowledge to iteratively implement, evaluate, and customize the technology for the individual [20, 51, 63]. This theme extends the literature by highlighting that during a transition period in an adult with IDD's life, the knowledge of a well established caregiver may be lost and the consequences of this are dire as it leads to regression in skills as well as a subsequent loss of independence. This highlights a key challenge for assistive technology that is introduced at the point of transition: it starts from scratch, without the accumulated history that makes caregiving so effective in addressing hidden needs.

## 4.4 Theme 4: Participants' Perceptions and Experiences with Assistive Technology

Theme 4 wraps up our findings by examining parents' perspectives on assistive technology. Parents reflected on where they saw technology playing a meaningful role in their adult child's life and what alternative approaches they believed could better address the challenges they faced.

*4.4.1 Where Technology Should Be Used.* Families of individuals with limited communication ability in both verbal and reading comprehension (P11, P8A/B, P7) had greater exposure to assistive technologies than their counterparts. Yet notably, all participants expressed a consistently positive outlook on the role of assistive technology in supporting growth and agency. As P5 puts it, *"I think all of this has the potential to help him do something that he can't otherwise do."* P6 emphasized how assistive technology could be transformative for agency and self-determination: *"I think he's always pretty proud when he's able to accomplish something on his own, and he should have that opportunity as much as possible. [...] I think technology affords that."* Technology was perceived as an enabler of access, autonomy, and possibility.

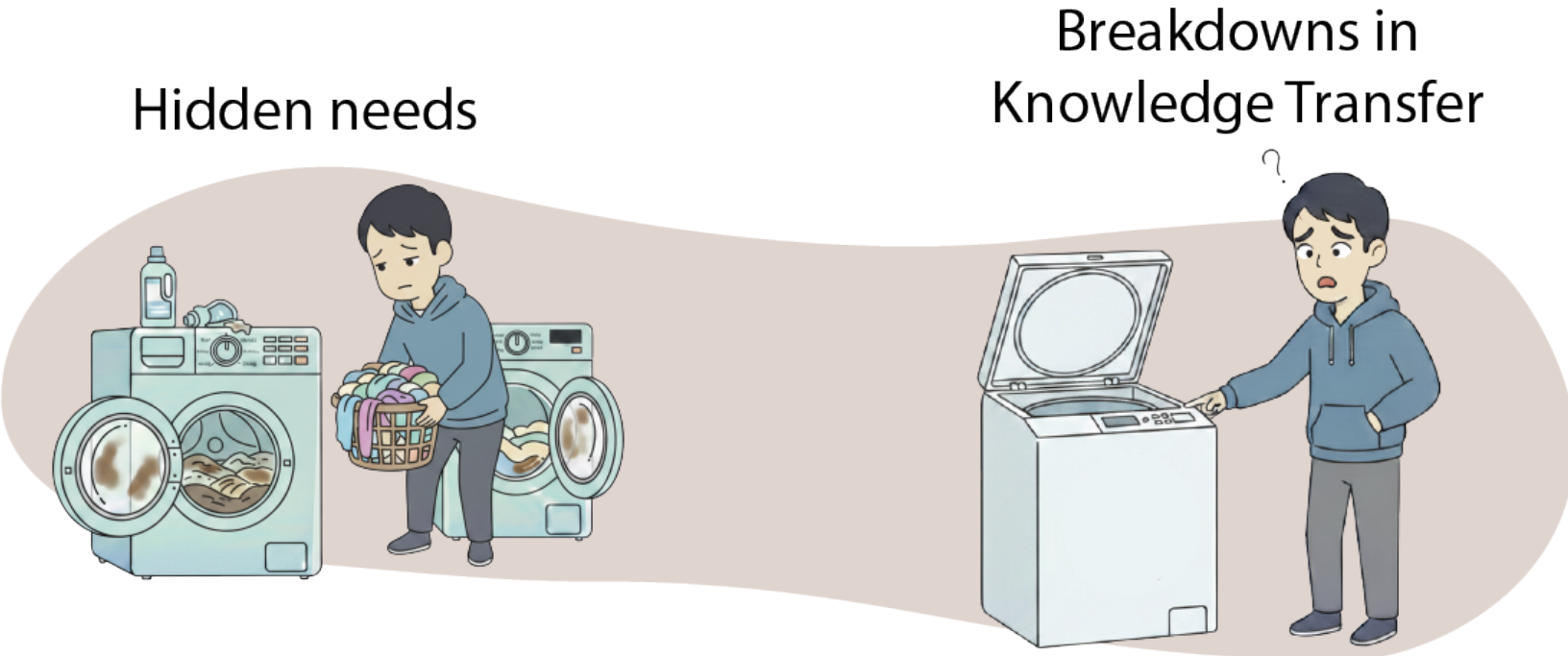

Figure 4: Transition challenges across environments and caregivers, illustrated through laundry. Left: a hidden need, where the adult with IDD does not vocally express that the detergent is out and continues the routine without it. Right: a breakdown in knowledge transfer, where a new machine causes the adult with IDD to struggle with a task they had already mastered at home.

This optimism is paired with clear expectations of technology. Participants emphasized that technology should complement, not replace, human presence in moments where physical touch and genuine connection are essential (e.g., illness, loss, or emotional need). Similarly, P5 believes, *"while technology can enrich daily experiences and increase exposure to beauty, it should not substitute real-world activities such as visiting a museum."* Participants articulated boundaries around where and how technology should be used. Many preferred it limited to shared spaces (e.g., common areas and kitchens), with private spaces like bedrooms and bathrooms off-limits. There were also concerns about over monitoring, which may make individuals feel they are under constant scrutiny. Privacy and security were non-negotiable. Participants expected technology to protect adults with IDD from exploitation, safeguard personal data, and prevent unwanted intrusion into their homes.

*4.4.2 Familiar Shortcomings, New Possibilities.* Families noted that in several areas existing assistive technologies fell short of their needs or envisioned potential. These include the lack of customizability and intuitiveness; fragmentation across single-use apps; and becoming antiquated and losing perceived value. Solutions meant to reduce cognitive burden of activities of daily living created new learning curves. Promising technologies can quickly be ruled out when they lack flexibility or present undesirable interfaces. P10, whose daughter experiences sudden falls triggered from unexpected stimuli, described the difficulty of finding appropriate ways for technology to prompt:

> *"I would love for her to have like an apple watch, but [...] they beep or they vibrate, and that won't work for us. [...] It would be so cool to have something that would kind of prod her to get stuff done, but it would make her fall."*

Even when customization was possible, it was often burdensome, as P11 detailed her experience with a visual schedule planner app: *"I'm manually [inputting] every single step of every single task."* She expressed how automation could make this seamless: *"But there's got to be a way of [...] electronically set up from a, you know, a logical set of systems. Could you take a video, put the video in, and the app breaks it down into steps?"*

P8A noted how the cohesive UI and seamless syncing of Apple devices made life easier for him as a neurotypical user, but he wished there was an equivalent for adults with IDD: *"[if] he doesn't have to go to a different device or think about how to operate those in a different way, because the UX is different. It comes to one spot, and all that complexity is hidden through the integrations [...] everything looks and feels and works the same way, regardless of the task that you're trying to to achieve. That would be brilliant."*

This inconsistency is compounded for support staff, who manage more than one adult with IDD at a time. When every device is different, managing across them becomes difficult, and solutions are more likely to be abandoned. P9 described this breakdown in continuity: *"Since he finished school at age 26, he has not had anybody who has consistently interacted with those kinds of programs or schedules or [devices] with him when he's not with us. So it tends to fall by the wayside."*

*4.4.3 Design Insights.* The positive attitudes toward assistive technology documented in literature hold true in our findings, as do the criteria that literature calls for it to meet [33, 57]. Aligning with existing literature, parents approached assistive technology with the same optimism that it could potentially help but re-emphasized that this optimism is contingent on privacy, ease of use, and customizability to an individual with IDD's ability. Our findings extend literature as it highlights that individuals with IDD transitioning to a different living situation challenged by fragmentation in interfaces across technology ecosystems and motivation of stakeholders within an individual's care network, which in-turn hinders adoption across settings. This lack of consistency will not only hinder growth but worse yet has the potential to cause regression in an individual with IDD.

# 5 Discussion

This study asked how parents provide support, how they perceive technology, and what assistive technology would need to succeed in a new home. Themes 1 and 2 address RQ1, showing how parents selected and structured support to match their adult child's capacities and foster growth over time. Theme 4 addresses RQ2, highlighting how existing technologies fall short of this standard, often requiring the person to adapt to the tool. Theme 3 addresses RQ3, pointing to the challenge of carrying knowledge forward to maintain continuity of support across the transition to semi-independent living. Our findings make it clear that parental support extends beyond task assistance to encompass broader developmental goals for their adult children. This discussion examines what that means for assistive technology design.

## 5.1 Interpretation of Findings

*5.1.1 Support Targets Developmental Goals.* Parents described three modes of support: task adaptation, self-adaptation via routines, and doing-for. Their clear preference was to lead with approaches that enabled growth and agency. Doing-for appeared when safety demanded it or when time pressure made it unavoidable. Parents framed their core goal as helping their adult child build and maintain skills so they can grow toward greater independence.

Notably, the decision-making logic parents used when providing support aligns with frameworks validated across multiple clinical disciplines including occupational therapy and special education. These frameworks have different terminology such as habilitation and systematic instruction but the same underlying goal of building an individual's skill over time [4, 8, 16, 75]. Parents described many of the evidenced-based practices these fields utilize such as graduated prompting, feedback fading, skill chaining, and tracking of competence over time. Across all fields, the same underlying decision-making logic is followed where support should be calibrated to current ability, delivered at the edge of what the individual can do, and faded as competence grows.

*5.1.2 A Mismatch in Design Logic.* Assistive technology has largely been designed for compensatory support that helps individuals manage what they cannot do, rather than support growth toward what they could [49]. Zubatiy et al. frame conversational agents as compensatory technologies that fill gaps as cognitive decline gradually reduces individuals' with mild cognitive impairment (MCI) ability to participate in a task over time [80]. Other work with MCI further show this pattern where systems have been designed to help older adults with MCI complete daily tasks safely and success is not measured by growth but by task completion and reduced caregiver burden [14, 62]. Assistive technology for adults with IDD largely inherits this model with reviews showing systems relying on fixed prompts and predefined routines, which prioritizes task completion over developmental goals [20, 34].

For adults with IDD, support should extend beyond the immediate goal of task completion and include the developmental trajectory over time. Work with blind and visually impaired children points toward this developmental trajectory as Gadiraju et al. found that when prompts were designed to encourage parents to scale back assistance, independence followed [25]. Adults with IDD approaching semi-independent living require assistive technology built on that same principle. While such systems currently do not exist for aiding daily living, similar technologies being developed for classroom settings such as knowledge tracing [1] could be adapted for supporting domestic learning and growth.

*5.1.3 The Knowledge Behind the Support.* The habilitative support parents described depended on knowledge that others did not have. Parents knew the prompt their adult child needed today versus yesterday, which step of a task was the right one to fade, and what a difficult morning meant for the rest of the day. This kind of knowing aligns with what Polanyi [55] called tacit knowledge: implicit, experience-based understanding that is difficult to articulate or codify. Research on caregiving for people with intellectual and multiple disabilities has shown that this knowledge is built through years of proximity and repeated interaction that plays a central role in caregiving [31, 41]. Even structured handover protocols in nursing consistently lose individualized knowledge between shifts [54], and in eldercare tacit knowledge has been described as the invisible layer of caregiving most at risk of being lost when care moves across people and settings [3]. At the transition to semi-independent living, adults with IDD face this loss of tacit knowledge. The following design principles address this gap by asking what it would take for assistive technology not only to deliver habilitation-oriented support, but to carry forward the knowledge that makes support effective in the first place.

## 5.2 Design Principles for Growth-Oriented Assistive Technology

We developed several design principles for how assistive technology can be improved to support continuity and growth in the transition to semi-independent living. These were formulated from direct suggestions in our interviews, and inferential analysis of gaps between parental supports and what assistive technology currently can do (Fig. 5 and Fig. 6).

*5.2.1 Build on Caregivers' Tacit Knowledge With Pre-Transition Onboarding.* The transition to semi-independent living risks disrupting the continuity of person-specific knowledge that made that support from a caregiver effective [31, 60]. Much of this knowledge is tacit. The design challenge is finding ways to capture and build from what caregivers know before that knowledge is lost in transition. This makes pre-transition onboarding especially important so that systems begin building from caregiver knowledge while parents are still present to demonstrate routines, explain support choices, and help contextualize what the system captures.

Pre-transition onboarding could include routines, communication styles, safety concerns, preferred prompts, and signs of difficulty that are easy to miss. Current assistive technology largely supports reminders, prompting, and step-by-step task guidance [20, 27, 44, 71], but it has not typically been designed to capture

**Design Principle: 5.2.1 Tacit Knowledge**

| Challenges | Design Recommendations |
|---|---|
| The Initiate-Assist-Verify (IAV) support structure relies predominantly on parents (Theme 2) | Design pre-transition onboarding that captures routines, preferred prompts, communication styles, safety concerns, and signs of difficulty that are easy to miss |
| Modifications to this structure are identified manually, based on knowledge gained through years of experience caring for the adult with IDD (Theme 2) | Capture why a support choice works and how it should shift across contexts |
| Skills regress when knowledge isn't effectively transferred across contexts and caregivers (Theme 3) | Establish individual baseline to enable anomaly detection over time |
| Unexpressed symptoms and issues can have severe consequences (Theme 3) | |

**Design Principle: 5.2.2 Foster Long Term Growth**

| Challenges | Design Recommendations |
|---|---|
| Growth is tracked manually (Theme 2) | Follow IAV as an organizing structure for task support |
| Continuous calibration of the IAV structure is necessary (Theme 2) | Expand success metrics beyond task completion to include growth outcomes |
| There is no mechanism for technology to adjust support as competence develops (Theme 2) | Evolve IAV structure to reflect adults with IDD's growth and regression over time |
| Parental overinvolvement can unintentionally undermine agency and growth (Theme 1) | Avoid flagging every deviation from the expected path as a mistake |
| | Calibrate support when sustained participation is a more appropriate goal than growth |

Figure 5: Summary of design principles: tacit knowledge and fostering long-term growth. Each card defines a core principle, the challenges identified in the findings, and the resulting design recommendations. Mappings are not one-to-one, and each recommendation may address multiple challenges.

why a support choice works for one person, how it should shift across contexts, or what should count as typical for that individual over time. Starting earlier would not remove the difficulty of transferring tacit knowledge, but it could help preserve more of the support logic that might otherwise be lost during transition.

*5.2.2 Foster Long Term Growth. Where Growth Is Not Possible, Promote Participation and Maintenance.* Parents described support as a structure that we formulate here as IAV. Importantly, this structure was not fixed. Parents adjusted support as the individual's capacity changed over time by fading help where growth was occurring and maintaining support where it was still needed. Assistive technology should be able to follow a similar logic but current assistive technologies for daily living are largely designed to help a person complete a task through reminders, prompts, and step-by-step guidance [20, 27, 44, 71]. Our findings show that this is not enough for the transition of adults with IDD to semi-independent living because effective support is not only about successful task execution but growth focused. This means it is also necessary to know when support should fade, when it should remain in place, and when the goal should shift from growth to continued participation. Growth-oriented assistive technology should therefore move beyond fixed prompting and follow an evolving IAV structure that reflects the individual's development over time. As P11 put it, *"I would love AI to go, 'Oh, great! He's learned those steps. Let's not give him those steps anymore. Let's get him to the harder level' without having to manually go in there, delete it and reconfigure it."*

Shifting some of this tracking from parents to technology also has implications for agency and motivation. Prior work shows that adults with IDD value independence, and that technology should be better designed to recognize ability, support progress, and reinforce what users can do, rather than frame assistance around paternalistic control [26, 33, 61]. At the same time, not every deviation should be treated as an error and technology shouldn't enforce a single correct path as it risks undermining growth and agency. In some cases, growth may not be the right goal and sustained participation and maintenance may be more appropriate outcomes. Assistive

**Design Principle: 5.2.3 Sense and Adapt**

**Challenges**

- Support needs fluctuate daily based on individual's internal state and external environmental variability (Theme 2)
- Current technology only delivers static prompts (Theme 2)
- Detecting subtle signals of difficulty or distress requires personal knowledge about the individual (Theme 3)

**Design Recommendations**

- Calibrate support based on internal state and environmental conditions
- Learn to recognize behavioral signals of difficulty
- Actively probe wellbeing rather than wait for self-report
- Balance adaptive sensing with user privacy

**Design Principle: 5.2.4 Individualized Customization with Single Entry Point for the Care Network**

**Challenges**

- Technologies fail to accommodate the diverse communicative styles and speech patterns (Theme 1)
- Configuration and maintenance is manual and tedious (Theme 4)
- Disparate and bespoke tools lead to fragmentation (Theme 4)
- High learning curves lead to abandonment (Theme 4)

**Design Recommendations**

- Prioritize low cognitive load, natural interactions, minimal user interface, and support multi-task usage
- Take a unified system approach and prioritize portability i.e. ensure consistent design conventions throughout various environmental contexts
- Design for parameterized customization

Figure 6: Summary of design principles: sense and adapt, and individualized customization with a single entry point for the care network. Each card defines a core principle, the challenges identified in the findings, and the resulting design recommendations. Mappings are not one-to-one, and each recommendation may address multiple challenges.

technology should therefore not only adapt support as skills develop, but also recognize when preserving participation is itself a meaningful success.

*5.2.3 Sense and Adapt, Rather Than Follow Scripted Routines.* Current widely available assistive technology provides static prompts and assistance but lack the sensing capabilities to verify or adapt their support to changing needs [20]. Even families who were heavy adopters of assistive technology had to re-prompt their adult child when reminders were snoozed and provide corrections where necessary.

Adaptive feedback is such a critical element of learning that we believe it will be vital for realizing the growth-oriented vision described above. Pilot studies have demonstrated how specialized smart homes can enhance independence through such adaptive feedback, but these trials relied on remote teleoperators [19] or bespoke sensors and logic [42, 45]. This presents barriers to customization and widespread adoption.

This need for adaptation has been noted in prior literature [20], but solutions remained elusive. Recent advances in video understanding are beginning to address some of this challenge [15] by enabling general perception of daily living activities. However, generalist internet-trained models lack the knowledge of individualized behaviors, patterns, and interventions that enable effective adaptive support. We suggest tacit knowledge may be a key component to solving this problem. This knowledge might be gleaned by combining recent advances in wearables and egocentric perception [21] to learn some of the patterns and internal and external variables described in Theme 2.2. However, we recognize the limitations of automatic perception. Passive data collection introduces complex privacy and ethical concerns, and no system can perfectly perceive focus, energy level, emotions, transitions, or comfort in different settings. Therefore, AI perception should be supplemented by explicit input where possible. Some existing technologies have been developed to help individuals with IDD articulate their mood and mental wellbeing [43, 73]; we suggest integrating this data to modulate their other supports. Navigating the complex ethical considerations of technologies that learn and adapt will require engagement of diverse stakeholders in co-design to avoid invasive paternalism.

*5.2.4 Individualized Customization with a Single Entry Point for the Care Network.* Throughout this paper, we emphasize the necessity of tailored systems for growth-oriented solutions. However, traditional customization remains difficult to sustain in practice. This is largely due to a paradox: technologies intended to reduce the cognitive burden of activities of daily living often introduce significant new learning curves [9]. Users are currently expected to learn the mechanics of a technology, adapt to specific input requirements, and navigate inherent limitations. As highlighted in our findings and aligning with literature, these burdens are compounded by the increasing involvement of external caregivers who often struggle with the fragmented and unintuitive interfaces of existing assistive applications [63]. As a result, current systems fail not only individuals with IDD, but also the broader care networks required to support long-term adoption, contributing to the limited or abandoned use documented in adjacent populations [70]. Current systems miss a critical combination: customization for the individual with IDD paired with a shared entry point that is intuitive, learnable, and usable across an entire care network of parents and rotating staff.

Participants demonstrated a preference for general-purpose technologies, such as voice assistants and smart displays, echoing prior findings [33], including work showing that adults with IDD readily adopt voice interfaces [5] and that support workers and adults with IDD can engage with the same voice-based system [50]. These systems are easy to adopt because they are designed so that with minimal training users can intuitively engage with the technology across multiple tasks through a single entry point (e.g., single unified interface) such as natural language commands. Namely, they prioritize low cognitive load, natural interactions, minimal user interfaces, and support multi-task usage.

We advocate that future assistive technology should leverage these concepts used in general-purpose technologies to improve the interaction, configuration, and personalization experience. Drawing on Theme 4, participants envisioned unified systems that: 1) replicate the interaction patterns of everyday technologies that provide a single entry point for care network interaction, and 2) support individualized assistance to an adult with IDD. A single entry point could take several forms: natural language input as a universal interaction layer, a consistent design language across tools (as in the Apple ecosystem), or a central gateway that routes users based on need rather than exposing underlying complexity (as in a search interface).

Rather than distributing functionality across many task-specific applications that have varying user interfaces, a single system should also support diverse abilities and activities through parameterized customization for adults with IDD. Visual, textual, and verbal supports could be tailored to individual adults with IDD while maintaining a consistent underlying interaction logic and navigation flow (i.e., single entry point), preserving familiarity for both caregivers and individuals with IDD, a balance that prior work identifies as both critical and difficult to achieve, since consistency is itself a core cognitive accessibility requirement [23] that dynamic adaptation can undermine [30].

User-specific interface adaptations have been historically difficult to scale because each adaptation had to be manually designed and configured for the individual. Recent advances in LLM-assisted UI adaptation are beginning to lower this barrier, with systems that modify existing interfaces [47] or generate new ones [56] to match a user's accessibility needs. These approaches could make parameterized customization feasible at the individual level while preserving the consistent interaction logic the care network depends on.

## 6 Limitations and Future Work

This study has several limitations. Our sample of parents from 13 families provides very diverse perspectives but are not necessarily generalizable to the entire community given the vast variation of needs and abilities present in individuals with IDD. Additionally, parent perspectives alone do not provide a complete picture. Including adults with IDD directly in the design process is essential to avoid paternalistic assumptions and ensure their preferences are reflected. We look forward to building on these findings through need-finding with the adults with IDD in the families we interviewed. A third limitation concerns how parental knowledge would be captured in practice. Our findings argue this knowledge is essential for habilitative support, but we do not address how it would be collected, who would own it, or how errors would be corrected. In future work, we plan to apply these findings to explore technology integrations in the home, investigate the tradeoffs and ethical nuances of monitoring, and engage in co-design of new assistive technologies with adults with IDD.

## Acknowledgments

(1) We thank our partner organization, whose work focuses on building neuro-inclusive residential communities, for supporting this research. We are especially grateful to the parents and caregivers who generously shared their time and experiences with us.
(2) We acknowledge the use of Google Nano Banana in supporting the creation of Figures 1–4.